\documentclass[10pt,twocolumn,letterpaper]{article}

\usepackage{iccv}
\usepackage{times}
\usepackage{epsfig}
\usepackage{graphicx}
\usepackage{amsmath}
\usepackage{amssymb}

\usepackage[breaklinks=true,bookmarks=false]{hyperref}

\iccvfinalcopy 

\def\iccvPaperID{****} 

\begin{document}

\title{Early Estimation of User's Intention of Tele-Operation \\Using
  Object Affordance and Hand Motion in a Dual First-Person Vision}

%

\author{Motoki Kojima and Jun Miura\\
Department of Computer Science and Engineering\\
Toyohashi University of Technology
}

\maketitle

\begin{abstract}

  This paper describes a method of estimating the intention of a user's
  motion in a robot tele-operation scenario. One of the issues in
  tele-operation is latency, which occurs due to various reasons such
  as a slow robot motion and a narrow communication channel. An
  effective way of reducing the latency is to estimate human intention
  of motions and to move the robot proactively. To enable a reliable
  early intention estimation, we use both hand motion and object
  affordances in a dual first-person vision (robot and user) with an
  HMD. Experimental results in an object pickup scenario show the
  effectiveness of the method.

\end{abstract}
\section{Introduction}
 \label{sec:intro}

Lifestyle support is one of the promising application domains of
robotic technologies. A mobile manipulator is a useful tool for the
user with a disability to perform various home tasks such as bringing
a user-specified object from a remote room. Tele-operation through a
robot's view is a common approach to easily controlling such a robot.
Human body motion-based control (e.g.,
\cite{HaICRA-2015})
is promising because it can
provide an intuitive way of tele-operation.  There are, however, two
issues in body motion-based control. One is the body motion
observation, and the other is the latency due to motion estimation or
robot control.

Concerning body motion estimation, existing devices such as MoCap,
RGB-D cameras, and wearable sensors are costly and/or not very
reliable. Concerning the latency, even if the communication channel is
fast enough, the latency is sometimes unavoidable due to, for example,
the difficulty in controlling a robot precisely. A hybrid approach
will then be a solution, which combines the user's command and robot's
autonomous movement based on object recognition and motion planning
capabilities, but keeping the intuitiveness of operation remains as a
problem.

\begin{figure*}[t]
\begin{center}
   \includegraphics[width=0.8\linewidth]{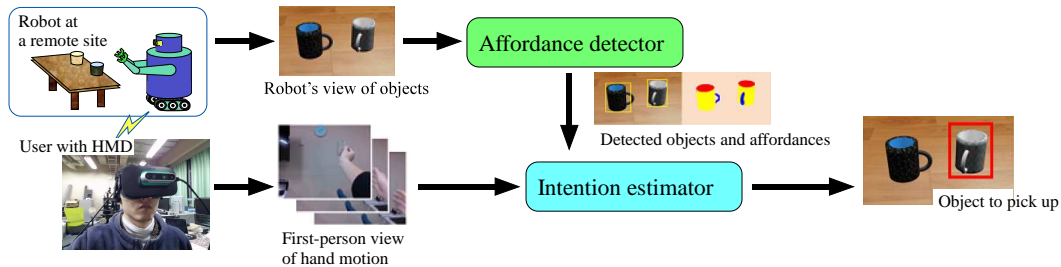}
   \caption{An intended scenario and the overview of the proposed
     action intention estimation method. The user with an HMD sees the
   image from the robot and trying to pick up one of the objects by moving
   his/her hand. The method analyzes both the first-person view of the
   hand and the robot view to determine which object he/she wants to
   pick up.}
   \label{fig:Overview}
\end{center}
\end{figure*}

To address these issues, we propose an approach based on early
intention estimation of human actions with a {\em dual} first-person
view of the robot and the user. We suppose the situation where a user
tries to pick up one of the objects on a remote table.
Fig. \ref{fig:Overview} shows an intended scenario and an overview of
the proposed method. The user obtains the robot's current view through
a head-mounted display (HMD). The user reaches out a hand to touch a
target object in the view. Of course, the hand never touches the
object at a remote site. The system uses a stream of hand images for
intention estimation. For an early but reliable estimation of
intention, we use a strong relationship between human hand shape
(i.e., preshaping \cite{BardIROS-1991}) and object affordances.
Unlike the previous work on action recognition in a first-person
vision, we combine data from different views.

\section{Related Work}
\label{sec:RelatedWork}

\subsection{Action recognition by first-person vision}

Many methods have been proposed for End-to-End action recognition by
first-person vision \cite{SinghCVPR-2016,RhinehartPAMI-2018}.  These
methods analyzes the relationships of objects and hand motions which
both appear in first-person image sequences.  In our tele-operation
scenario, these methods cannot be directly adopted because hands and
objects are at different locations and therefore observed in separate
images.

\subsection{Hand motion intention estimation using preshaping and
  affordances}

Characteristics of an object which naturally induces a human action is
called {\em affordance} \cite{Affordance}. Arbib et
al. \cite{Arbib-1985} show the relationship between the shape and the
motion of a hand and the visible characteristics of the target object.
This implies that preshaping provides useful information for
identifying the target object \cite{NakamuraIEICE-1997}. In our
problem setting, the user cannot get a visual feedback of her/his own
hand position and motion. Some work show that even in such a case, as
long as the location of the target of interaction is known, the
reaching motion can be executed similarly
\cite{WingesEBR-2003,Shimawaki-2011}; This supports the use of
separately observed object locations/types and a hand motion stream.

\section{Object Affordance Extraction}
\label{sec:Affordance}

Do et al. \cite{DoICRA-2018} developed {\em AffordanceNet}.
Affordance here is a part of an object which is closely related to
some human action. AffordanceNet outputs object names and bounding
boxes, and affordance-wise segmentation.

\begin{figure}[tb]
\begin{center}
   \includegraphics[width=0.9\linewidth]{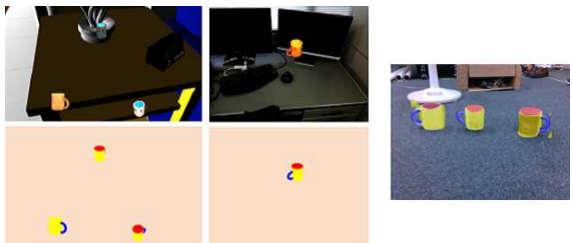}
   \caption{Example data with a simulated (left) and a real (center)
     background and an affordance extraction exmaple (right).}
   \label{fig:AffordanceNetDataPrediction}
\end{center}
\end{figure}

We use cups as target objects. We modified the original AffordanceNet
such that the class output is only ``cup'' and the affordance output
has the following four classes: {\em contain}, {\em wrap-grasp}, {\em
  handle-grasp}, and {\em background}. We also constructed a new
dataset accordingly by ourselves.
Fig. \ref{fig:AffordanceNetDataPrediction} shows examples in the new
dataset and an affordance extraction result.

\section{Intention Estimation Network}
\label{sec:LSTM}

\subsection{Network structure}

\begin{figure}[tb]
\begin{center}
   \includegraphics[width=\linewidth]{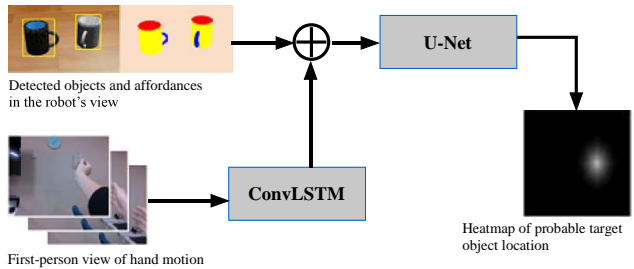}
   \caption{Intention estimation network.}
   \label{fig:IEN}
\end{center}
\end{figure}

Early intention estimation or action prediction usually necessitates
integration and analysis of time series data. We design a network,
called {\em Intention Estimation Network (IEN)}, which is to calculate
probable target location(s) of the current hand motion based on an
integration of affordance information and hand information (see
Fig. \ref{fig:IEN}). For the former, we use the output of
AffordanceNet trained with our dataset, which is composed of a set of
four affordance masks and a bounding box mask. For the latter, we use
the output of the ConvLSTM \cite{ConvLSTM} with a sequence of hand
images as an input.  We concatenate both outputs and then supply it to
a variant of U-Net \cite{RonnebergerArXiv-2015}, one of the popular
networks for semantic segmentation. The final heatmap is normalized so
that the values of all pixels sum to one.

\subsection{Dataset construction for intention estimation}

The input to IEN includes the extracted object and affordance
infomration as well as a hand image sequence. The latter is the
heatmap of the target location. Since the output of AffordanceNet
for a real setting is not free from errors, we first generated a
variety of simulated scenes, which are with ground truth information
of affordances and object bounding boxes, and then provided them to
the user for taking a hand-reaching motion.

Our system uses a stereoscopic head-mounted display (HMD) (Occulus
Rift CV1) for presenting images of the remote site to the user.  We
use an RGB-D camera (Intel RealSense D435) for acquiring hand motion
sequences, which are recorded both in RGB and depth images.
Fig. \ref{fig:Overview} shows a user wearing the system on the bottom
left.

\begin{figure}[tb]
\medskip
\begin{center}
   \includegraphics[width=0.6\linewidth]{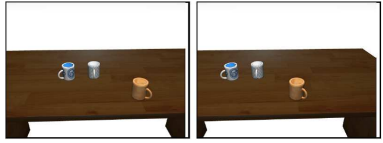}\\
   {\scriptsize (a) Stereo image pair. Grasping the handle of the orange cup is
     the target task.}\\
   \vspace*{2mm}
   \includegraphics[width=0.85\linewidth]{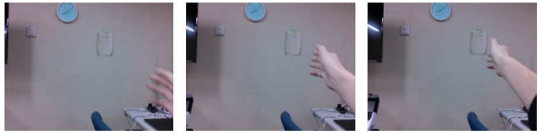}\\
   {\scriptsize (b) Hand motion sequence.}
   \vspace*{1mm}
   \caption{An example pair of stereo image and hand motion.}
   \label{fig:HandMotionExample}
\end{center}
\end{figure}

The steps for collecting a hand motion sequence for an affordance/stereo
pair are as follows (see Fig. \ref{fig:HandMotionExample}):
\begin{enumerate}
  \setlength{\columnsep}{-3mm}\setlength{\itemsep}{-1mm}
  \item
    Put two or three cups on the table. Textures,
    orientations, and locations of cups are set randomly.
  \item
    A stereo image pair is shown on the HMD.
  \item
    The user moves the right hand to try to pick up the designated object in
  the scene, and the hand image sequence is recorded with the RGB-D camera.
  \item
    The recorded sequence is saved with the output of AffordanceNet. 
\end{enumerate}

\section{Performance Evaluation}
\label{sec:Evaluation}

\subsection{Training and testing the network}

We collected 156 pairs of affordance and hand motion information for
training. One hand motion sequence consists of 60 frames (30fps, 2sec)
and a set of 10-frame sequences are extracted with a sliding-window
fashion. As a result, we have 50 sequences for each original sequence
and 7,800 sequences in total.  The ground truth of the output heatmap
is generated by putting a Gaussian distribution at the location of the
designated object. We used the KL-divergence as the loss function to
evaluate the distance between the true and the predicted heatmap.

We generated an additional test dataset for evaluation by 50 trials.
The input to the intention estimation network (IEN) in testing is not
limited to a fixed-length (10 frames) sequence but a set of sequences
with variable lengths from 1 to 15 to see how early the intention can
be estimated.

We examined the following three points: (1) effectiveness of
affordance information, (2) comparison of RGB and depth images, and
(3) effectiveness of hand region extraction in RGB images. 

\subsection{Estimation examples}

\begin{figure}[tb]
\begin{center}
   \includegraphics[width=0.85\linewidth]{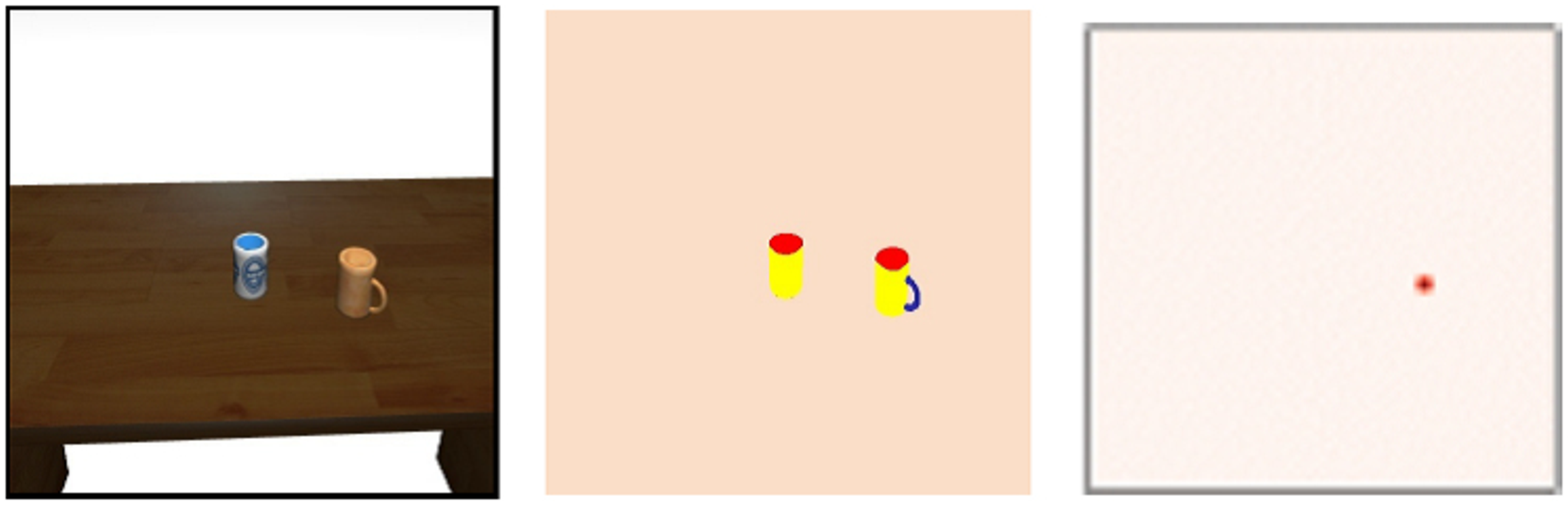}\\
   {\scriptsize Test scene, extracted affordances, and the ground truth heatmap.}\\
   \vspace*{1mm}
   \includegraphics[width=0.85\linewidth]{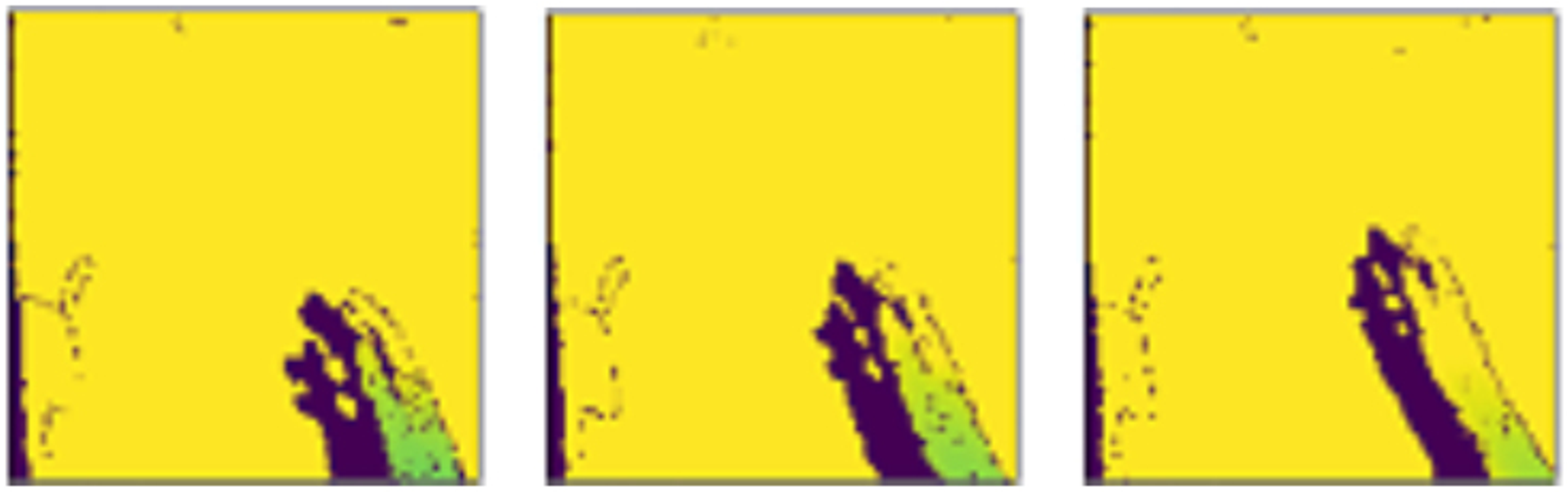}\\
   {\scriptsize Input depth images at 2nd, 4th, and 6th frame (from left to right).}\\
   \vspace*{1mm}
   \includegraphics[width=0.85\linewidth]{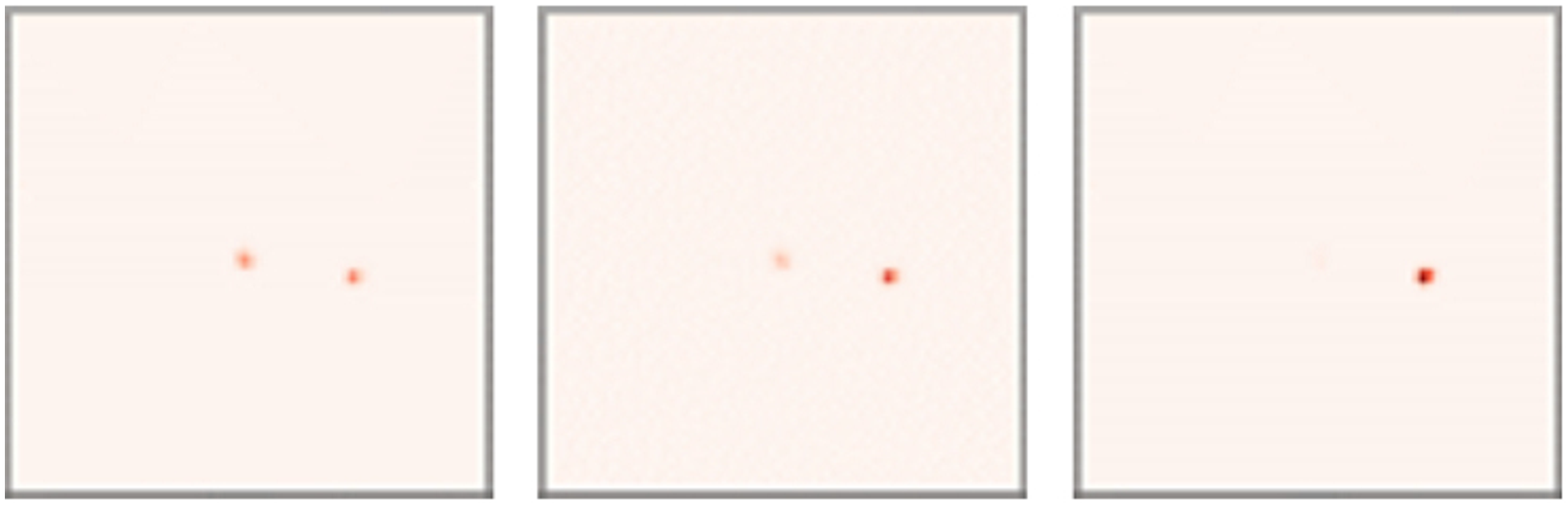}\\
   {\scriptsize Estimation results with affordance.}\\
   \vspace*{1mm}
   \includegraphics[width=0.85\linewidth]{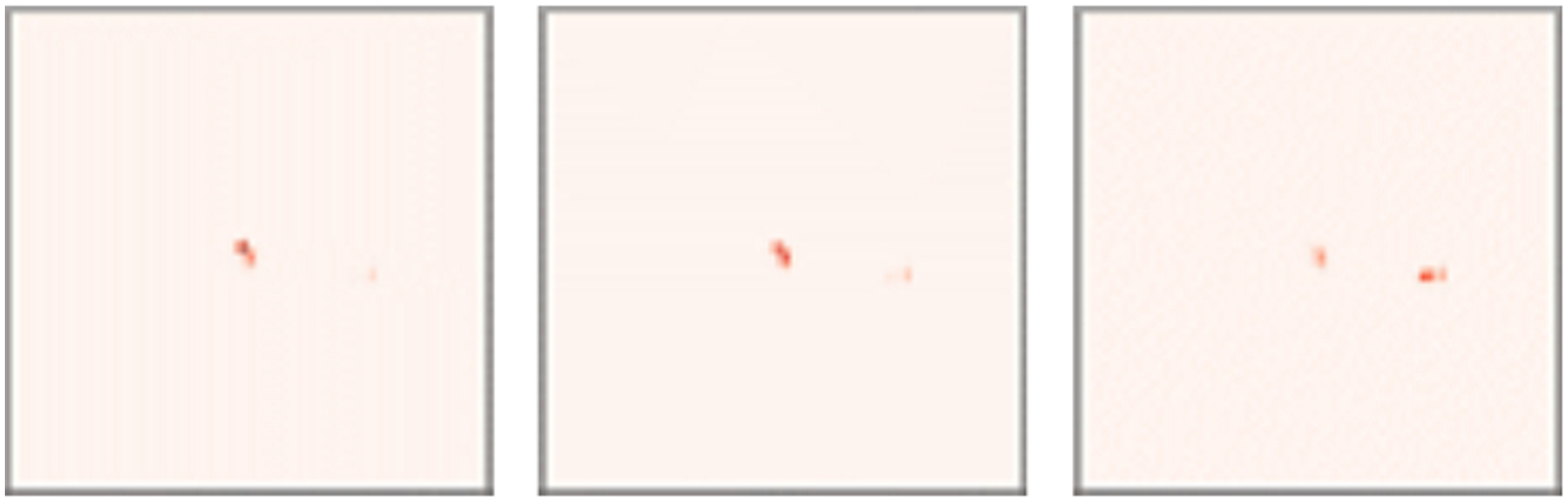}\\
   {\scriptsize Estimation results without affordance.}\\
   \vspace*{1mm}
   \caption{Intention estimation results for depth images.}
   \label{fig:Case1}
\end{center}
\end{figure}

Fig. \ref{fig:Case1} shows the inputs and the corresponding results
for depth images. There are two cups, and the user is asked to pick up
the right one. With affordance information, the probability of the
right cup gradually increased, while without affordance information,
the probabilities for both objects remain almost the same.

\begin{figure}[tb]
\begin{center}
   \includegraphics[width=0.85\linewidth]{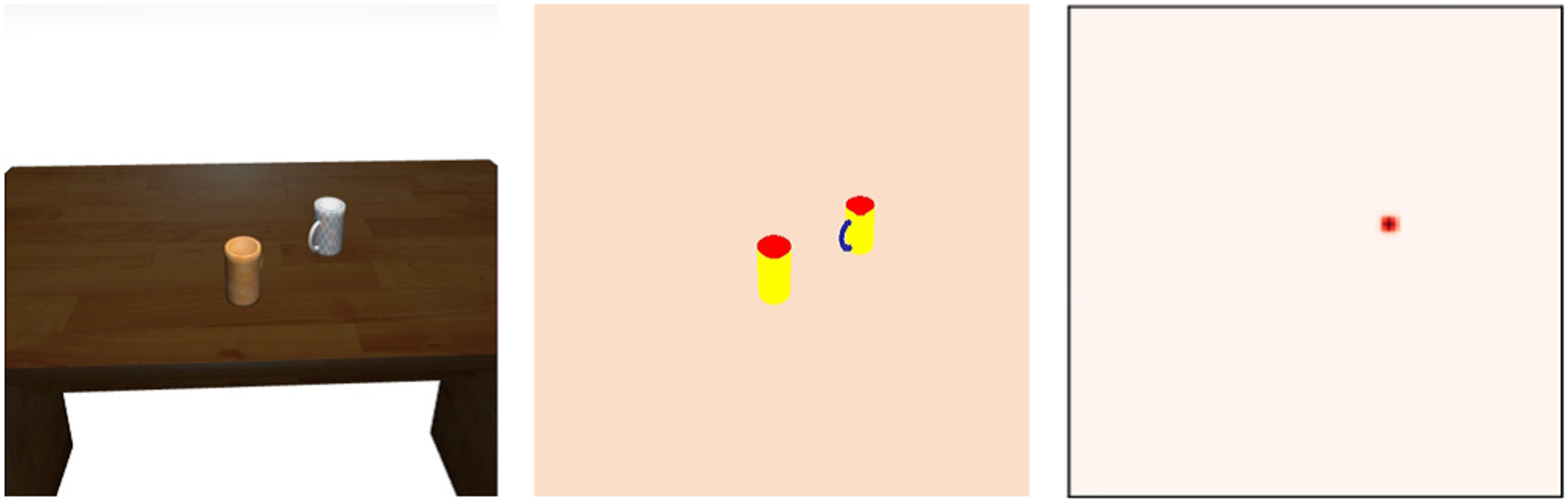}\\
   {\scriptsize Test scene, extracted affordances, and the ground truth heatmap.}\\
   \vspace*{1mm}
   \includegraphics[width=0.85\linewidth]{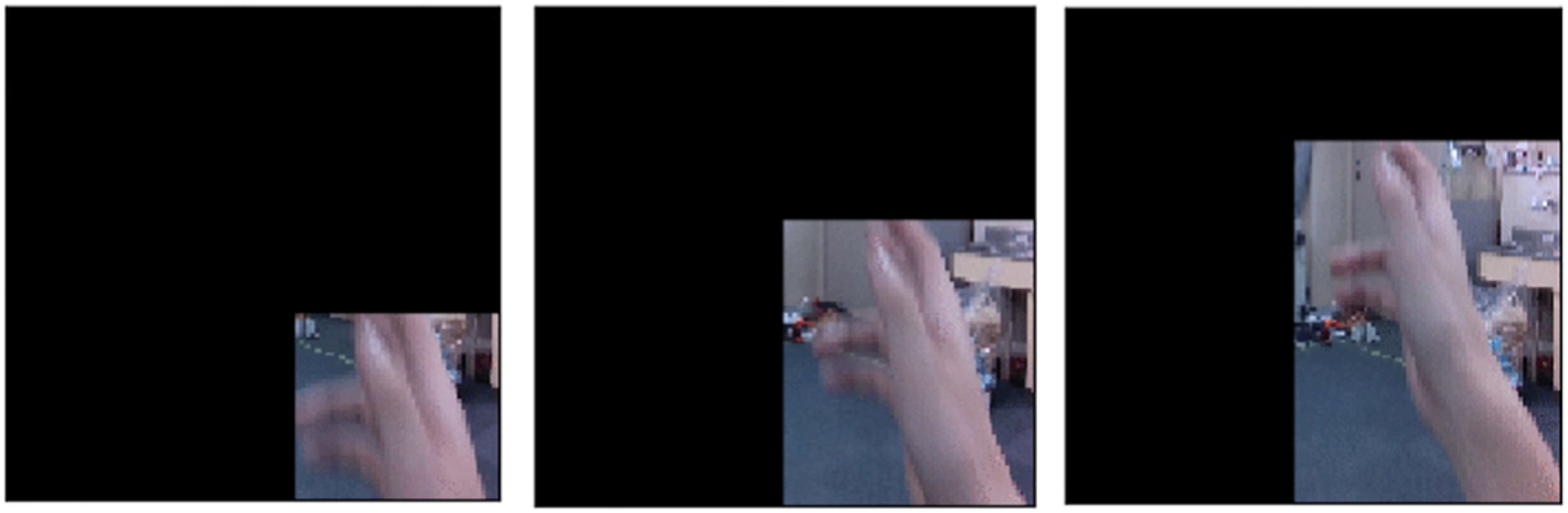}\\
   {\scriptsize RGB images with hand extraction at 2nd, 4th, and 6th frame.}\\
   \vspace*{1mm}
   \includegraphics[width=0.85\linewidth]{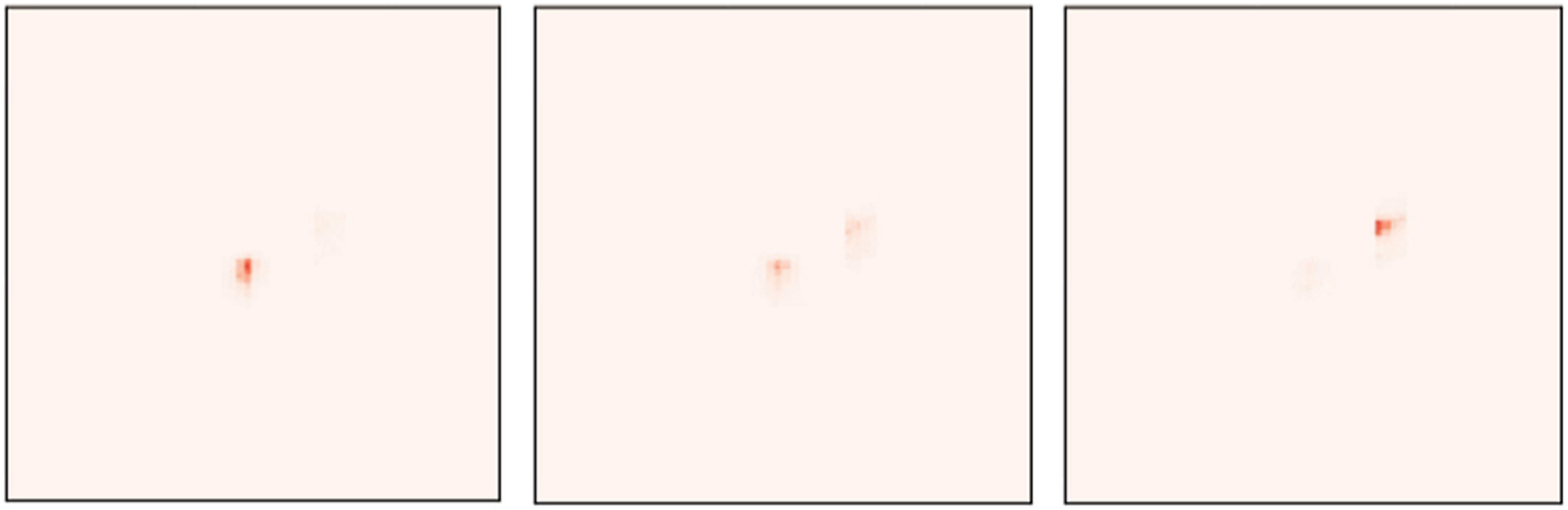}\\
   {\scriptsize Estimation results with hand extraction and affordance.}\\
   \vspace*{1mm}
   \includegraphics[width=0.85\linewidth]{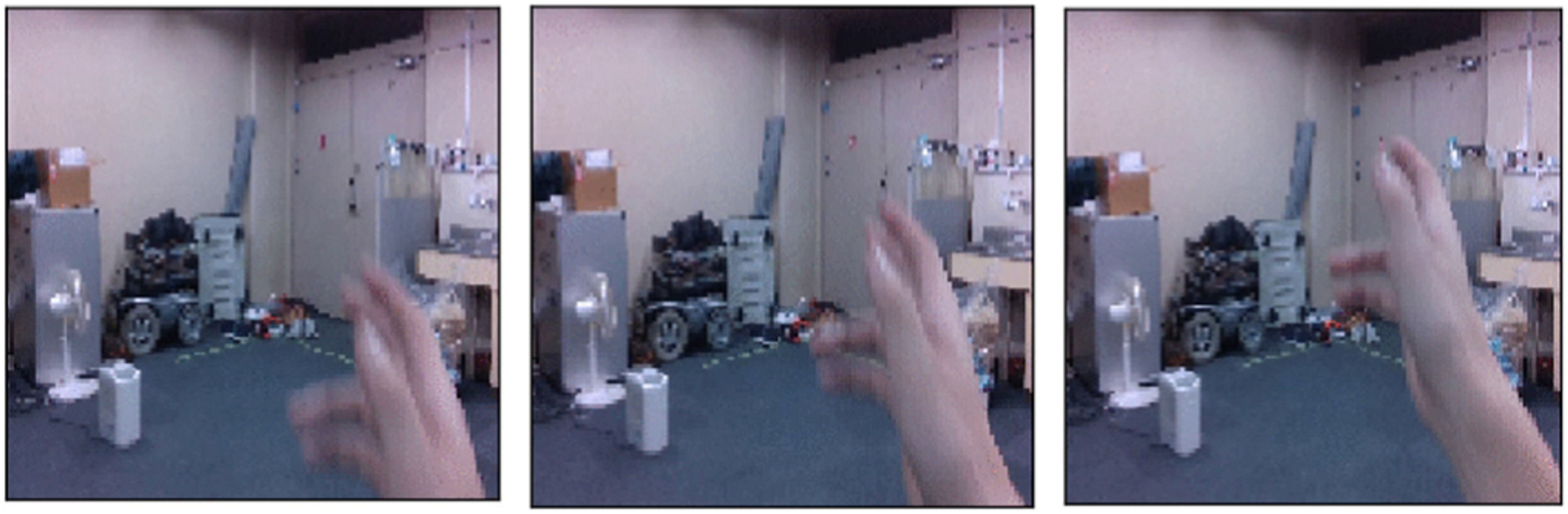}\\
   {\scriptsize Input RGB images at 2nd, 4th, and 6th frame.}\\
   \vspace*{1mm}
   \includegraphics[width=0.85\linewidth]{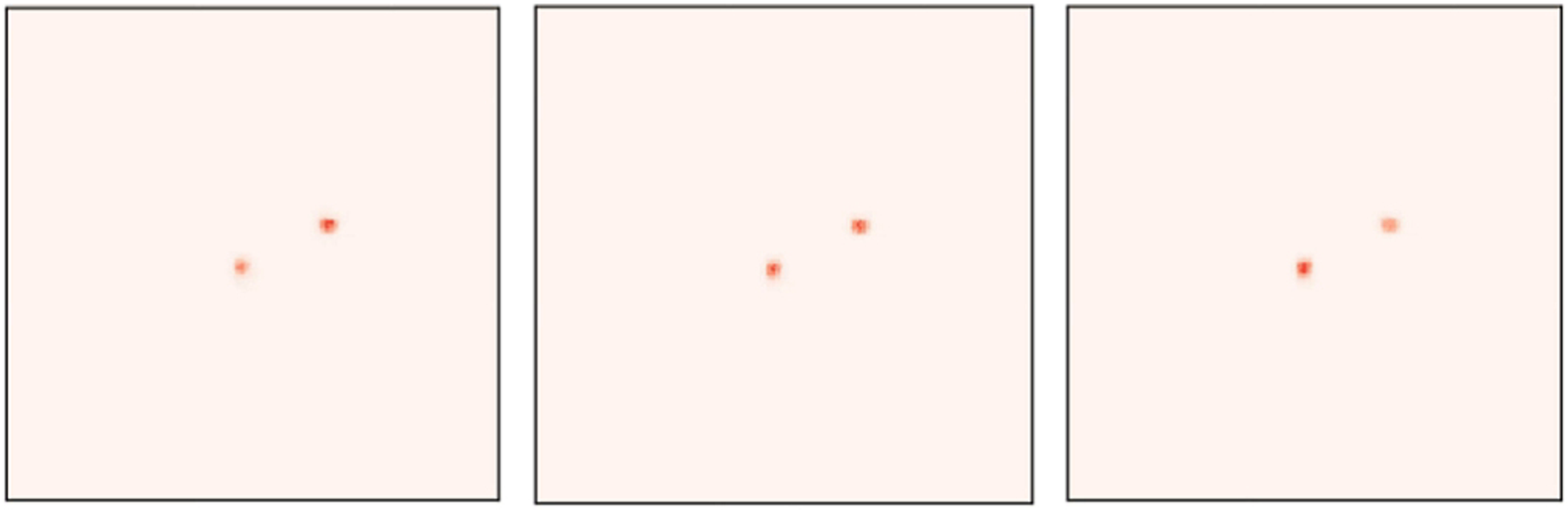}\\
   {\scriptsize Estimation results with affordance but without hand
     extraction.}\\
   \vspace*{1mm}
   \caption{Intention estimation results for RGB images.}
   \label{fig:CaseR}
\end{center}
\end{figure}

Fig. \ref{fig:CaseR} examines the usefulness of hand extraction for
RGB images. There are two cups, and the user is asked to pick up the
right one. With hand region information, the probability of the right
cup increased at the 6th frame, while both cups are equally probable
without that information. Since we learned the use of hand extraction
is always useful, we include hand extraction for RGB images in the
subsequent quantitative comparison.

\subsection{Quantitative evaluation}

The purpose of the intention estimation is to determine the target
object as early as possible. Since a heatmap can be interpreted as a
probability distribution of the location of the target object, we use
the sum of probabilities with the bounding box of an object as the
confidence value. Such confidence values for all objects in the scene
are normalized to be used as probabilities of being the target.
The object whose probability exceeds a threhold $th_{target}$ is
judged as the intended target object. We use the averaged f-value as a
measure of estimation quality. 

Table \ref{tab:Comparison} compares actual f-values for four cases in
order to summarize the above comparisons. Using depth images is more
effective in the intention estimation because the hand region and
shape information is more clearly obtained compared to RGB
images. Using affordance information is useful for both depth and RGB
images, and using RGB images with hand extraction and affordances is
comparable to using depth images without affordances.

\begin{table}[t]
  \begin{minipage}{\linewidth}
    \small
\begin{center}
   \caption{Summary of comparison in terms of f-values.}
   \label{tab:Comparison}
   $th_{target}=0.6$\\
\begin{tabular}{|c||c|c|c|c|c|c|}
  \hline
   case/frame & 2 & 4 & 6 & 8 & 10 & 12\\
  \hline\hline
  Depth-AO &{\bf 0.405} &{\bf 0.487} &{\bf 0.695} & {\bf 0.745} &{\bf
    0.761} &{\bf 0.725}\\
  \hline
  Depth-O &0.038 &0.466 &0.600 &0.610 &0.625 &0.658\\
  \hline
  RGB-AO &0.242 &0.286 &0.350 &0.432 &0.568 &0.667\\
  \hline
  RGB-O &0.039 &0.145 &0.317 &0.286 &0.556 &0.691\\
 \hline
\end{tabular}\\
\vspace*{1mm}
   $th_{target}=0.8$\\
\begin{tabular}{|c||c|c|c|c|c|c|}
  \hline
   case/frame & 2 & 4 & 6 & 8 & 10 & 12\\
  \hline\hline
  Depth-AO &{\bf 0.175} &{\bf 0.323} &{\bf 0.507} & {\bf 0.615} & {\bf
    0.675} &{\bf 0.707}\\
  \hline
  Depth-O &0.039 &0.113 &0.323 &0.355 &0.500 &0.556\\
  \hline
  RGB-AO &0.038 &0.172 &0.197 &0.338 &0.493 &0.640\\
  \hline
  RGB-O &0.000 &0.039 &0.113 &0.182 &0.276 &0.381\\
 \hline
\end{tabular}\\
{\scriptsize *-AO: with affordance, *-O: without affordance}
\end{center}
 \end{minipage}
\end{table}

\section{Summary}
\label{sec:Conclusion}

This paper proposes a method for early estimation of the user's
intention of tele-operation using a dual first-person images
and object affordances. We deal with a tele-operated object pickup
scenario where the user sees a view of a remote site through an HMD
and moves his/her hand to try to pick up a target
object. AffordanceNet extracts object and affordance information and
ConvLSTM extracts hand motion and shape information. The U-Net
combines these two kinds of information to generate a heatmap of the
target location. We generated two new datasets, one is for training
AffordanceNet and the other is for training the whole network. We
compared several cases and show that affordance information is useful
in early intention estimation both for depth and RGB images. Testing
the developed method on a real robot and applying it to other objects
are future work.

\subsection*{Acknowledgment}

This work is in part supported by JSPS KAKENHI Grant
Numbers 17H01799. The authors would like to thank Mr. Hiroaki Masuzawa
for fruitful discussion.


\end{document}